# Performance Analysis of Deep Autoencoder and NCA Dimensionality Reduction Techniques with KNN, ENN and SVM Classifiers

Md. Abu Bakr Siddique[1], Shadman Sakib[2], Md. Abdur Rahman[3]
[1,2]Deprtment of EEE, International University of Business Agriculture and Technology, Dhaka-1230, Bangladesh
[3]Department of EEE, Islamic University of Technology, Gazipur-1704, Bangladesh
absiddique@iubat.edu[1], bbkrsddque@gmail.com[1], sakibshadman15@gmail.com[2], marahman.iut@gmail.com[3]

*Abstract*—The central aim of this paper is to implement Deep Autoencoder and Neighborhood Components Analysis (NCA) dimensionality reduction methods in Matlab and to observe the application of these algorithms on nine unlike datasets from UCI machine learning repository. These datasets are CNAE9, Movement Libras, Pima Indians diabetes, Parkinsons, Knowledge, Segmentation, Seeds, Mammographic Masses, and Ionosphere. First of all, the dimension of these datasets has been reduced to fifty percent of their original dimension by selecting and extracting the most relevant and appropriate features or attributes using Deep Autoencoder and NCA dimensionality reduction techniques. Afterward, each dataset is classified applying K-Nearest Neighbors (KNN), Extended Nearest Neighbors (ENN) and Support Vector Machine (SVM) classification algorithms. All classification algorithms are developed in the Matlab environment. In each classification, the training test data ratio is always set to ninety percent: ten percent. Upon classification, variation between accuracies is observed and analyzed to find the degree of compatibility of each dimensionality reduction technique with each classifier and to evaluate each classifier performance on each dataset.

*Keywords—Dimensionality Reduction Techniques, Deep Autoencoder, Neighborhood Components Analysis (NCA), Feature selection, Feature extraction, UCI machine learning repository, Supervised classification*

I. INTRODUCTION

In machine learning, statistics and data science problems, especially in data classification and regression problems, dimensionality reduction is the method of truncating the number of features or attributes under examination by picking up a set of prime features or attributes [1]. This method can be implemented by both feature selection as well as feature extraction [2]. Feature selection is the process of eliminating irrelevant and redundant features from the existing data-features which do not have any bearing on class identification [3]. Contrarily, feature extraction evaluates the whole dataset and maps the data contents into a lower-dimensional feature space from high dimensional feature space.

The importance of dimensionality reduction is profound in big-data analysis, regression and classification problems as bigger data require lower representation. As in recent times, the datasets are overloaded with countless features; the chance of overfitting is very high as the model gets progressively more reliant on the data it is trained on. The major motivation behind the implementation of dimensionality reduction techniques is to reduce and eliminate overfitting. Dimensionality reduction of dataset not only reduces overfitting of the model but also ensures enhance performances, faster training and less cost, less storage spaces requirement for data, removal of noise and extraneous features, operation of algorithms unsuitable in high dimensions, and so on.

In unsupervised feature learning for single modalities, deep networks have been effectively applied like in content, pictures or sound and so on. A deep autoencoder [4] is a feed-forward neural network comprising of an input layer, an output layer and at least single or multiple hidden layers. It comprises of two principal parts: an encoder network which compresses the n-dimensions of the input dataset into an m-dimensional space and a decoder networks which enlarges the vector data from an m-dimensional space into the main n-dimensional dataset and takes the data back to their distinctive values. It is normally utilized for nonlinear dimensionality reduction, for a set of data via training the network to disregard the noise in the signal. Deep autoencoders are viably utilized for resolving many applied issues, from face recognition to obtain the semantic meaning for the words [5].

Another very effective dimensionality reduction technique is Neighborhood Component Analysis (NCA). This non-parametric method is used to find feature extraction by maximizing the stochastic deviant of leave-one-out nearest neighbor gain to achieve the best accuracy. NCA performs dimensionality reduction by learning a k*d matrix rather than d*d matrix where d represents the total number of original high dimensions and k represents the number of new lower dimensions of picking.

In this paper, we have developed the Matlab modeling of Deep Autoencoder and NCA dimensionality reduction algorithms as well as KNN, SVM and ENN classification algorithms. We have chosen nine special datasets from the UCI machine learning repository for feature extraction and classification purposes. These datasets are CNAE9, Segmentation, Movement Libras, Knowledge, Pima Indians diabetes, Parkinsons, Seeds, Mammographic Masses, and Ionosphere. Firstly, the dimensionality of each dataset is reduced by 50% of the original dataset by selecting the relevant and major half features or attributes from each dataset using Deep Autoencoder and NCA dimensionality reduction methods. To evaluate the performance of each dimensionality method, we implemented KNN, ENN and SVM classification algorithms in Matlab. Firstly, these algorithms are trained on training sets and then their performances are assessed on validation sets. Finally, from the comparison of the observed variations of the accuracies, the performance of each algorithm can be evaluated. The steps involved in the whole process are given below as a block diagram.

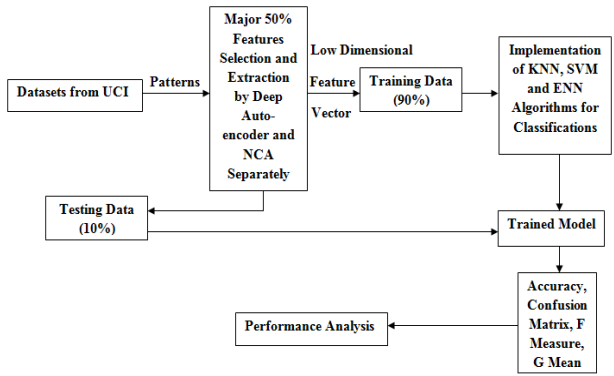

Fig. 1. Block diagram signifying the whole process in the paper

## II. LITERATURE REVIEW

Pattern classification algorithms are generally used to manage high-dimensional data. Dimensionality reduction techniques undertake a significant part in pattern identification problems where the feature vectors mainly place on a high-dimensional feature space. It is challenging to legitimately employ machine learning algorithms to this sort of task due to the invented curse of dimensionality. Deep autoencoders (DAE) [6] including additional deep neural networks have exhibited their viability in finding non-linear features through over numerous problem areas. DAE limits the reformation inaccuracy of the bona fide input to acquire a low dimensional depiction which precisely gives a nonlinear expansion of PCA. DAE has been utilized to adapt low-dimensional depictions, and a deep neural system for protecting the group neighborhood makeup [7]. Zhuang et al. [8] utilized a supervised interpretation learning technique for transfer learning dependent on deep autoencoders. Autoencoded dynamic development primitive is a model proposed by Chan et al. [9] which uses DAE to discover a description of development in a latent feature space. For the further improvement of their model, they included sparsity in the feature layers of the neurons so that different changes can be viewed noticeably. A general nonlinear implanting system was dependent on a deep neural system proposed by Huang et al. [10] which can be used to execute a group of dimensionality reduction algorithms. On the other hand, a few nearest-neighbor based algorithms have been created during the past few years, for example, neighborhood component analysis (NCA) [11] is one of them. NCA is an algorithm that uses a method alike k-nearest-neighbors to find out a space in which neighborhoods of point having similar names are closer than focuses with various names. Nearest-neighbor is a straightforward and productive nonlinear decision rule and often yields modest results differentiated and state-of-the-art classification techniques, for example, support vector machines and neural networks. NCA endeavors to evaluate a linear makeover by amplifying the stochastic variant of the usual KNN gain on the training set. An unsupervised NCA algorithm proposed by Qin et al. [12] for clustering which can become familiar with a low-dimensional estimation of high dimensional data accordingly can seal as an unsupervised dimensionality reduction. In addition, a new language identification framework [13] is proposed through I vectors. The framework is assessed based on various DR techniques and NCA is one of them.

## III. OVERVIEW OF ALGORITHMS

### A. Deep Autoencoder

Normally, when an autoencoder contains a number of hidden layers it is called deep autoencoder. The mission of the neural network in autoencoder is to reproduce its very own input subject to the limitation that one of its hidden layers is of low-dimension. By allowing numerous layers of encoders and decoders, a deep autoencoder can successfully symbolize complex distributions over the input. To replicate the input vector against the output layer, the backpropagation algorithm is used to train the deep autoencoders [14]. Therefore to denote the input vector, it diminishes the numeric sections in the data by utilizing the output of the hidden layer.

The mathematical development of deep autoencoder with a hidden layer can be summed up by defining the encoder and decoder as $\alpha$, and $\hat{x}$.

For encoding,
$$\alpha = f(Wx+b) \quad \ldots\ldots\ldots (1)$$

For decoding,
$$\hat{x} = f'(W'\alpha + b') \quad \ldots\ldots\ldots (2)$$

Where $f$ and $f'$ are the nonlinear activation function, $W \in \mathbb{R}^{l \times m}$ and $W' \in \mathbb{R}^{m \times l}$ are the weight matrices, $b \in \mathbb{R}^l$ and $b' \in \mathbb{R}^m$ are the bias vectors, and $\alpha \in \mathbb{R}^l$ is the output of the hidden layer.

By giving a set of input $\{x_i\}_{i=1}^{n}$, the reconstruction error can be calculated as:

$$\sum_{i=1}^{n} \|\hat{x}_i - x_i\|^2 \quad \ldots\ldots\ldots (3)$$

### B. Neighborhood Components Analysis (NCA)

Neighborhood Component Analysis (NCA) is a supervised non-parametric method employed in metric learning, classification and dimensionality reduction. This learning algorithm straight away maximizes a stochastic variation of the leave-one-out KNN performance on the training data [15]. Again, it learns a low dimensional linear embedding of labeled information to facilitate data visualization and prompt categorization in high dimensions [11]. Furthermore, this algorithm makes no prior supposition regarding the form of the group distributions or the margins among them. In essence, NCA utilizes a training set of vectors as input which is given by $\{x_1, x_2, x_3, \ldots, x_N\}$ where $x_i \in R^m$ and a corresponding set of labels $\{y_1, y_2, y_3, \ldots, y_N\}$ where $y_i \in L$. NCA finds a projection matrix A of dimension $p \times m$ where $Q = A^T A$ and this matrix estimates the training vectors $x_i$ into a p dimensional space. This projection matrix A defines a Mahalanobis distance between any two nearest neighbors in the projected space and is calculated by:

$$d(x_i, x_j) = (Ax_i - Ax_j)^T (Ax_i - Ax_j) \quad \ldots\ldots (4)$$

Now, if we select p<m, then we can learn the lower-dimensional representation of the original high dimensional data vectors.

The aim of this algorithm is to find a distance matrix A that maximizes the performances of the nearest neighbor classifier on the test data. For the optimization criterion, NCA takes advantage of stochastic neighbor assignments rather than simply using k nearest neighbors. Precisely, each test point j has a probability of $P_{ij}$ of allocating its label to its neighbor i. This probability perishes as the distance between points i and j increases:

$$P_{ij} = \frac{\exp(-\|Ax_i - Ax_j\|^2)}{\sum_{k \neq i} \exp(-\|Ax_i - Ax_k\|^2)}, \quad P_{ij} = 0 \quad \ldots\ldots (5)$$

In this method, for a single point, all points in the training data are considered as its neighbors but having different probabilities. Again, the probability of a particular point being the neighbor of its own is always considered as zero and in this respect, NCA is regarded as a leave-one-out classification. This stochastic choosing procedure finds the probability $P_i$ that point i being correctly identified as:

$$P_i = \sum_{j \in C_i} P_{ij} \quad \ldots\ldots (6)$$

$$\text{and, } C_i = \{j | y_i = y_j\} \quad \ldots\ldots (7)$$

The final optimization function f(A) is expressed as the sum of the probabilities of maximum expected numbers classified as correctly and it is calculated as:

$$f(A) = \sum_i P_i$$

To optimize the matrix A, the gradient of the optimization function can be found as:

$$\frac{\partial f}{\partial A} = 2A \sum_i \left( P_i \sum_k P_{ik} X_{ik} X_{ik}^T - \sum_{j \in C_i} P_{ij} X_{ij} X_{ij}^T \right) \quad \ldots\ldots (8)$$

NCA takes advantage of different gradient-based optimizers to calculate A. As the cost function f(A) is not convex, special care is needed for the initialization of matrix A, in order to avoid local minima of the solution and to find global minima precisely. According to the observation of the authors of this algorithm, they never experienced any effects of over-fitting in this technique and the performance of the algorithm by no means degraded as the training data size increased.

*C. K-Nearest Neighbors*

K-nearest neighbor (KNN) [16] is a case-based idle learning and nonparametric classifier accustomed to forecast the categorization of a novel test point in a dataset where data points are divided into a number of categories. KNN is supervised learning composed of a specified labeled dataset accommodating training sets (c, d) and like to signify the correlation between c and d. The objective of KNN is to find out a function $z: c \rightarrow d$ hence for a fresh test point c, z(c) can assertively deduce the equivalent output d. In KNN categorization, a novel test sample is categorized by a superior quantity of votes of its neighbors, with the test sample being allotted to the category mainly available amid its k nearest neighbors. The accomplishment of the KNN classifier is mostly hinged on the distance metric exercised to recognize the k nearest neighbors of a test sample. The most routinely employed one is the Euclidean metric represented by:

$$m(c,c') = \sqrt{(c_1 - c_1')^2 + (c_2 - c_2')^2 + \ldots + (c_n - c_n')^2} \quad \ldots\ldots (9)$$

For a specified quantity of nearest neighbors, k, and an unidentified test sample, c, and a distance metric m, a KNN algorithm evaluates conditional probability for every category. Finally, the unidentified test sample c is allotted to the category with maximum probability.

$$P(d = j | C = c) = \frac{1}{K} \sum_{i \in U} I(d^{(i)} = j) \quad \ldots\ldots (10)$$

Figure 2 illustrates the fundamental diagram of KNN classifier.

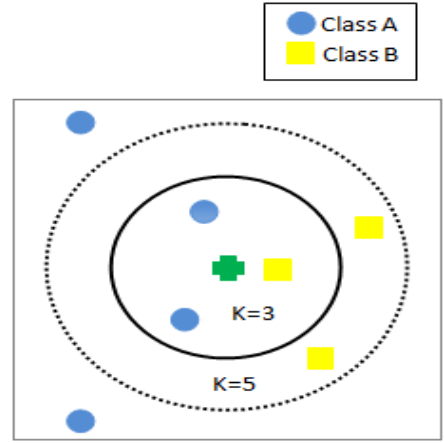

Fig. 2. Fundamental design of KNN classifier. If k = 3 the test sample (green polygon) is allotted to the category of blue circle and if k = 5 then it is assigned to the category of yellow square

*D. Extended Nearest Neighbors*

ENN categorizes a novel training point derived from the maximum gain of intra-class coherence. Contrasting KNN, wherein only the closest neighbors of a given test point are assessed for categorization, ENN categorizes a novel test point by not simply taking into account who are the next-door neighbors of the particular test point, rather also who regard as the test points as their bordering neighbors [17]. ENN utilizes the comprehensive category wise information from the entire training data to be trained from total data allotment and hence boosts categorization performance [18]. Let the comprehensive category style statistic $T_i$ for category $i$ concerning the collective examples $E_1$ and $E_2$ for every category next to its nearest neighbors can be calculated as:

$$T_i = \frac{1}{n_i k} \sum_{y \in E_i} \sum_{r=1}^{k} I_r(y, E = E_1 \cup E_2) \; ; i = 1, 2 \quad \ldots\ldots (11)$$

The indicator function $I_r(y, E)$ decides whether both the sample y as well as its $r^{th}$ nearest neighbors relate to the same group or not, can be evaluated as:

$$Ir(y,E) = \begin{cases} 1, & \text{if } y \in E_i \text{ and } NN_r(y,E) \in E_i \\ 0, & \text{otherwise} \end{cases} \quad \ldots\ldots (12)$$

Since Ti entitles the data allocation over several categories, the intraclass coherence can be shown as eqn. (13):

$$\Theta^j = \sum_{i=1}^{2} T_i^j \quad \ldots\ldots (13)$$

To categorize an unidentified sample Z, it is iteratively allocated to categories 1 and 2 correspondingly to achieve two novel comprehensive category style statistics $T_i^j$

$$T_i^j = \frac{1}{n_i'k} \sum_{y \in E'_{i,j}} \sum_{r=1}^{k} I_r\left(y, E' = E_1 \cup E_2 \cup \{Z\}\right) \quad ; i,j = 1,2 \quad \ldots\ldots (14)$$

Then the ENN classifies the sample Z based on the objective function as following:

$$f_{ENN} = \arg\max_{j \in 1,2} \sum_{i=1}^{2} T_i^j = \arg\max_{j \in 1,2} \Theta^j \quad \ldots\ldots (15)$$

The two category ENN categorization scheme can be simply expanded to multi-class categorization by:

$$f_{ENN} = \arg\max_{j \in 1,2,\ldots,N} \sum_{i=1}^{N} T_i^j \quad \ldots\ldots (16)$$

To steer clear of the recalculation of comprehensive category wise statistics $T_i^j$ in the testing period, the sample Z is iteratively allocated to every potential category j, j=1, 2, …., N, and regard the category involvement in accordance with a corresponding objective function of eqn. (14), expressed as:

$$f_{ENN.V} = \arg\max_{j \in 1,2,\ldots,N} \left\{ \left(\frac{\Delta n_i^j + k_i - kT_i}{(n_i+1)k}\right)_{i=j} - \sum_{i \neq j}^{N} \frac{\Delta n_i^j}{n_i k} \right\} \quad \ldots\ldots (17)$$

*E. Support Vector Machine*

The purpose of SVM is to create a framework established on training sample points to forecast the category labels of each test data specified only test data attributes. SVM classifier executes linear classification by achieving the hyperplane that enlarges the border between two categories. The data points that decide the hyperplane are called support vectors. In summary, for a set of specified labeled training data, the SVM classifier attains a most advantageous hyperplane which categories novel test data. Fig. 3 explains the essential design of the SVM classifier.

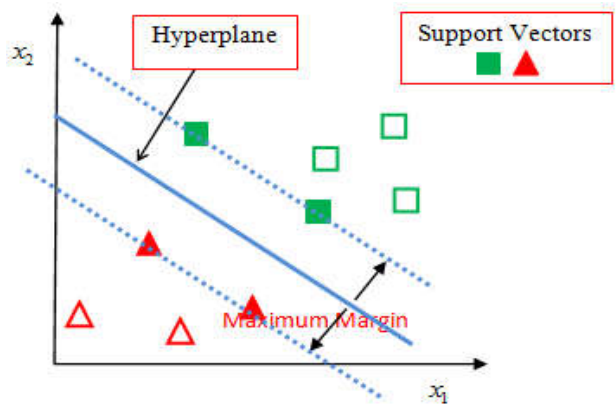

Fig. 3. The essential design of SVM classifier presenting the optimum margin, best possible hyperplane, plus the support vectors

For a specified training sample of n individual points with structure $(\vec{k}_1, l_1), \ldots\ldots, (\vec{k}_n, l_n)$ where $l_i = \pm 1$ representing the category affiliation of the bunch of points $\vec{k}_i$. A hyperplane can be showed by the cluster of points $\vec{k}_i$ as:

$$\vec{w} \cdot \vec{k}_i + a = 0 \quad \ldots\ldots (18)$$

To find the best hyperplane that divides the bunch of points $\vec{k}_i$ into either $l_i = +1$ or $-1$, the partition between the hyperplane as well as the nearby point $\vec{k}_i$ from each category should be optimized. Hence, the optimal hyperplane can be expressed as:

$$\vec{w} \cdot \vec{k}_i + a = \pm 1 \quad \ldots\ldots (19)$$

The distance between a sample data point with hyperplane can be shown as:

$$d = \frac{|\vec{w} \cdot \vec{k}_i + a|}{\|\vec{w}\|} \quad \ldots\ldots (20)$$

Now by setting the numerator to 1 from eqn. (19) in eqn. (20), the vacuity to the support vectors from a hyperplane is determined by:

$$d_{\text{Support Vectors}} = \frac{1}{\|\vec{w}\|} \quad \ldots\ldots (21)$$

Since hyperplane is expressed by a two-class example for $l_i = \pm 1$, the margin M is double the distance toward the adjoining instances:

$$\therefore M = \frac{2}{\|\vec{w}\|} \quad \ldots\ldots (22)$$

Lastly, the problem of maximizing M is equivalent to the problem of knocking off $\|\vec{w}\|$. To resolve this difficulty, the SVM classifier [19] entails the result of the following optimization problem:

$$\min_{w,a,\xi} \frac{1}{2} w \cdot w + C \sum_{i=1}^{n} \xi_i \quad \ldots\ldots\ldots\ldots (23)$$

with constraints: (i) $l_i(w \cdot \varnothing(k_i) + a) \geq 1 - \xi_i$
(ii) $\xi_i \geq 0$

The constrained optimization experiment of eqn. (23) can be transformed into an unconstrained optimization experiment via the Lagrangian function as expressed as eqn. (24): [19].

$$L(w,a,\xi) = \frac{1}{2} w \cdot w + C \sum_{i=1}^{n} \xi_i - \sum_{i=1}^{n} \alpha_i [l_i(w \varnothing(k_i) + a) - 1 + \xi_i] - \sum_{i=1}^{n} r_i \xi_i \quad \ldots\ldots (24)$$

Following the optimization of eqn. (24), the categorization assessment of a novel test data Z can be evaluated by utilizing the sign of eqn. (25). If the sign is (+)ve, the sample Z is in category 1 and if the sign is (-)ve Z is in category 2.

$$D(\vec{Z}) = \text{sgn}\left(\sum_{j=1}^{n} \alpha_j l_j \varnothing(\vec{k}_j) \cdot \vec{Z} + a\right) \quad \ldots\ldots (25)$$

IV. RESULTS AND DISCUSSION

In this study, we have used nine datasets (CNAE9, Knowledge, Seeds, Movement Libras, Pima Indians diabetes, Parkinsons, Segmentation, Mammographic Masses, and Ionosphere) in order to investigate the performance comparison among KNN, ENN and SVM after implementing two-dimensionality reduction techniques: Deep autoencoder and NCA. Figure 4 displays the classification accuracies for those datasets if Deep autoencoder is implemented as a dimensionality reduction technique.

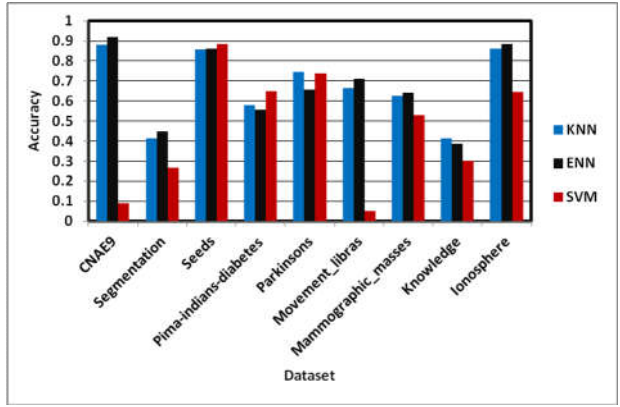

Fig. 4. Accuracy plot of KNN, ENN, and SVM for nine datasets using Deep Autoencoder

It is evident that, regarding classification accuracy, nearest-neighbor based algorithms outperform the SVM for the datasets in case of Deep autoencoder. For CNAE9, the ENN classifier demonstrated the highest accuracy (0.9194). On some datasets (e.g., Seeds, Pima-indians-diabetes, Knowledge) after using Deep autoencoder, SVM and KNN yield improvements over ENN. KNN showed the best performance for the CNAE9 dataset with an accuracy value of 0.8822. Nevertheless, the performance of the SVM classifier is inconsistent among the datasets with accuracy as low as 0.0495.

Table 1 displays the F measure and G mean values for the datasets using three different classifiers (KNN, ENN, and SVM) augmented with Deep autoencoder as a dimensionality reduction technique.

TABLE I. F MEASURE AND G MEAN WITH DEEP AUTOENCODER

| Dataset | F measure | | | G mean | | |
|---|---|---|---|---|---|---|
| | KNN | ENN | SVM | KNN | ENN | SVM |
| CNAE9 | 0.8772 | 0.9162 | 0.0404 | 0.8818 | 0.9184 | 0.0561 |
| Segmentation | 0.3913 | 0.4355 | 0.2001 | 0.4080 | 0.4470 | 0.2203 |
| Seeds | 0.8520 | 0.8534 | 0.8797 | 0.8573 | 0.8592 | 0.8840 |
| Pima-indians-diabetes | 0.5612 | 0.5490 | 0.3930 | 0.5671 | 0.5598 | 0.4025 |
| Parkinsons | 0.5814 | 0.5738 | 0.4225 | 0.5997 | 0.5825 | 0.4283 |
| Movement_libras | 0.6583 | 0.6998 | 0.0207 | 0.6731 | 0.7092 | 0.0308 |
| Mammographic_masses | 0.6240 | 0.6375 | 0.3454 | 0.6251 | 0.6387 | 0.3633 |
| Knowledge | 0.3683 | 0.3666 | 0.1167 | 0.3773 | 0.3776 | 0.1383 |
| Ionosphere | 0.8311 | 0.8706 | 0.3904 | 0.8424 | 0.8739 | 0.4005 |

On the other hand, Table 2 displays the corresponding information in case of NCA.

TABLE II. F MEASURE AND G MEAN WITH NCA

| Dataset | F measure | | | G mean | | |
|---|---|---|---|---|---|---|
| | KNN | ENN | SVM | KNN | ENN | SVM |
| CNAE9 | 0.6494 | 0.7110 | 0.8730 | 0.6571 | 0.7177 | 0.8780 |
| Segmentation | 0.7850 | 0.8042 | 0.8412 | 0.7942 | 0.8101 | 0.8514 |
| Seeds | 0.8602 | 0.8662 | 0.9408 | 0.8655 | 0.8713 | 0.9436 |
| Pima-indians-diabetes | 0.6883 | 0.7048 | 0.6859 | 0.6898 | 0.7084 | 0.6924 |
| Parkinsons | 0.7699 | 0.7456 | 0.6915 | 0.7824 | 0.7549 | 0.7091 |
| Movement_libras | 0.4573 | 0.4854 | 0.6816 | 0.4743 | 0.4958 | 0.6931 |
| Mammographic_masses | 0.7651 | 0.7818 | 0.7767 | 0.7663 | 0.7829 | 0.7798 |
| Knowledge | 0.8982 | 0.9088 | 0.9185 | 0.9017 | 0.9119 | 0.9212 |
| Ionosphere | 0.8557 | 0.8771 | 0.7832 | 0.8644 | 0.8794 | 0.7956 |

To all datasets for using Deep autoencoder and NCA we have reduced dimensionality to halves. Both the F measure and G mean values suggest that SVM demonstrates better performance if the NCA method is adapted. Contrariwise, after using Deep autoencoder, the ENN has an upper hand over KNN and SVM.

The performance evaluation based on the accuracy measurement among the classifiers is presented in Figure 5 where the NCA is playing the role as a dimensionality reduction technique.

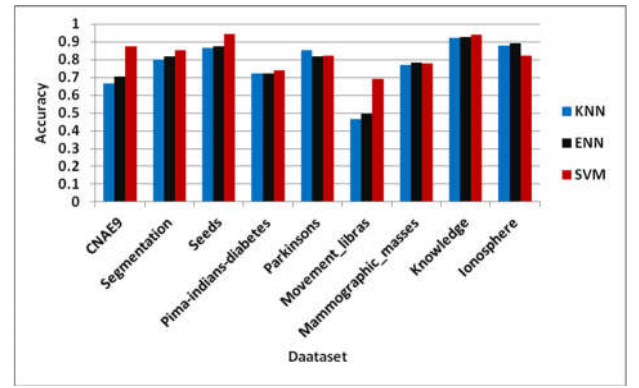

Fig. 5. Accuracy plot of KNN, ENN, and SVM for nine datasets using NCA

As observed from the Table 2, the aggregated performance of the SVM is better than KNN and ENN. The highest classification accuracy was witnessed in the case of the Seeds dataset (0.9452). However, in the case of some datasets (e.g., Parkinsons, Mammographic-masses, Ionosphere) the KNN and ENN outperform the Support vector machine classification algorithm.

On the other hand, the KNN classifier's performance generally hinges on the value of the k, the number of nearest neighbors selected regarding the dataset. In this study, the KNN classifier has shown remarkable performance for k=5. For some datasets, KNN classifier performs better than ENN after adapting the Deep autoencoder algorithm. The observations suggest that the Deep autoencoder has better compatibility with nearest-neighbor based algorithms; however, NCA is more compatible with the SVM technique.

## V. CONCLUSION

This paper has demonstrated the performance of Deep Autoencoder and NCA dimensionality reduction methods with KNN, ENN and SVM classifiers on nine different datasets from the UCI machine learning repository. From accuracy observation, it is clear that in the case of Deep Autoencoder, nearest-neighbor based classifiers have better performance on datasets than SVM classifier. ENN with Deep Autoencoder exhibited the highest accuracy of 91.94% for the CNAE9 dataset. In some datasets, although SVM and KNN demonstrated better performance than ENN, the performance of SVM is inconsistent among datasets with the lowest accuracy of 4.95% in the Movement Libras dataset. Now when the NCA dimensionality reduction technique is adapted, SVM generally outperforms both KNN and ENN classifiers and exhibited the highest classification accuracy of 94.52% for the Seeds dataset. It is observed from F Measure and G Mean values that if NCA is applied, SVM generally outperformed ENN and KNN. On the other hand, ENN works better with Deep Autoencoder than KNN and SVM. In conclusion, it can be stated that Deep Autoencoder is better compatible with KNN and ENN classifiers, whereas NCA is better compatible with the SVM classification method.